\documentclass[11pt]{article}

\usepackage[final]{acl}

\usepackage{times}
\usepackage{latexsym}

\usepackage[T1]{fontenc}

\usepackage[utf8]{inputenc}

\usepackage{microtype}

\usepackage{inconsolata}

\usepackage{graphicx}

\usepackage{amssymb}
\usepackage{booktabs}
\usepackage{multirow}
\usepackage{colortbl}
\usepackage{enumitem}
\usepackage[most]{tcolorbox}
\usepackage{adjustbox}
\usepackage{cleveref}

\definecolor{Highlight}{HTML}{39b54a}  
\definecolor{Cerulean}{HTML}{00A2E3}  
\newcommand{\hl}[1]{\textcolor{Highlight}{\bf #1}}

\newcommand{\PAR}[1]{\noindent {\bf #1.~}} 

\title{Beyond Bilingual Transfer: \\ Multilingual Code-Switching in Instruction Tuning}

\author{%
Shunta Asano\thanks{Equal contribution.} \quad
Jeonghun Baek\footnotemark[1] \quad
Toshihiko Yamasaki \\
The University of Tokyo \\
\texttt{asano@cvm.t.u-tokyo.ac.jp} 
}

\begin{document}
\maketitle

\begin{abstract}
Recent studies have shown that code-switching data (CSD), in which multiple languages are mixed within the same context, can improve cross-lingual transfer and multilingual alignment in large language models (LLMs). 
Recent studies on synthetic code-switching for LLMs have primarily focused on bilingual transfer between English and a target language, leaving instruction-tuning settings that jointly mix three or more languages within each example largely unexplored.
In this work, we investigate multilingual code-switching instruction tuning across four languages: English, Japanese, Korean, and Chinese. 
We evaluate multilingual understanding on Belebele and MMLU.
Our experiments show that simple sentence-level multilingual CSD consistently improves the four-language average accuracy across all bilingual, trilingual, and quadrilingual settings for Gemma-2-2B, indicating that multilingual code-switching can be effective beyond bilingual transfer settings. 
\end{abstract}

\section{Introduction}
Recent large language models (LLMs), including GPT-4~\cite{achiam2023gpt}, Gemini~\cite{comanici2025gemini}, Llama 3~\cite{grattafiori2024llama}, and Qwen3~\cite{yang2025qwen3}, have demonstrated strong multilingual capabilities across a wide range of languages. 
However, while English data is abundant in pretraining corpora, many non-English languages remain relatively low-resource, often leading to inconsistent performance across languages~\cite{yoo2025cscl, wang2025syncs}.

To address this challenge, recent studies have explored code-switching data (CSD), where multiple languages are mixed within the same context, to improve cross-lingual transfer and multilingual alignment~\cite{yoo2025cscl, wang2025syncs}. 
Prior approaches such as CSCL~\cite{yoo2025cscl} and SynCS~\cite{wang2025syncs} demonstrated that synthetic CSD effectively improves bilingual transfer between English and a target language. 
However, existing LLM studies primarily focus on bilingual settings involving only two languages, such as English-Korean or English-Chinese transfer. 
As a result, it remains unclear whether CSD remains effective during LLM instruction tuning when three or more languages are jointly mixed within the same training example.

Moreover, prior work mainly investigates CSD during pretraining settings~\cite{yoo2025cscl, wang2025syncs}. 
However, instruction tuning also plays an important role in multilingual capabilities~\cite{wei2022finetuned, ouyang2022training, chung2024scaling}. 
Despite this, multilingual CSD during instruction tuning remains underexplored.

\begin{figure*}[t]
    \centering
    \includegraphics[width=\linewidth]{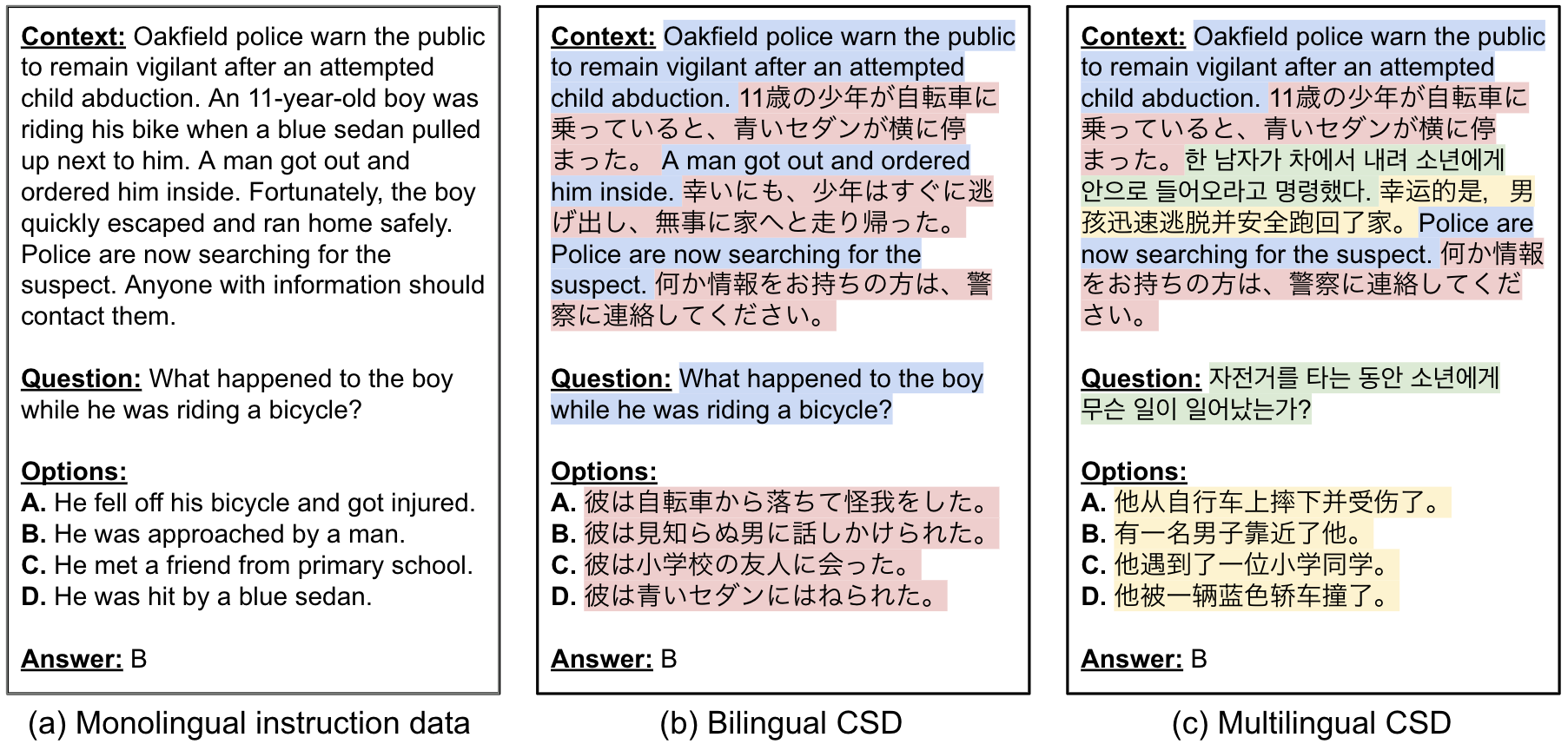}
    \caption{
    Illustration of (a) monolingual instruction data, (b) bilingual sentence-level code-switching data (CSD), and (c) multilingual sentence-level CSD used in our work. 
    Sentence-level CSD is constructed by combining sentence units from parallel multilingual instruction examples. 
    Despite its simple construction process, sentence-level CSD remains effective for multilingual instruction tuning.
    In CSD examples, different background colors indicate different languages.
    }
    \label{fig:csd_example}
\end{figure*}

In this work, we investigate multilingual code-switching instruction tuning across four languages: English, Japanese, Korean, and Chinese. 
We extend prior bilingual CSD approaches to multilingual settings and study whether CSD improves multilingual performance during instruction tuning.
To evaluate multilingual understanding, we conduct experiments on the multilingual multiple-choice benchmarks Belebele~\cite{bandarkar2024belebele} and MMLU~\cite{hendrycks2021measuring}.

For Gemma-2-2B, multilingual CSD consistently outperforms simple multilingual concatenation in average performance across bilingual, trilingual, and quadrilingual instruction-tuning settings on Belebele, with improvements also observed in trilingual and quadrilingual settings on MMLU.
These results suggest that multilingual code-switching remains effective even in more complex multilingual settings beyond pairwise bilingual transfer.
Our contributions are twofold. 
First, we extend synthetic code-switching during LLM instruction tuning beyond bilingual language pairs by jointly mixing three or more languages within the same example.
Second, we show that simple sentence-level CSD is effective for multilingual instruction tuning across four languages.

\section{Related Work}
Recent studies have explored code-switching data (CSD) to improve multilingual alignment and cross-lingual transfer in large language models (LLMs)~\cite{winata2023decades,yoo2025cscl,wang2025syncs}. 
Prior work showed that multilingual language models can implicitly learn cross-lingual alignment through shared multilingual representations~\cite{pires2019multilingual}. 
Recent studies further investigated synthetic code-switching, where multiple languages are mixed within the same context, to improve multilingual transfer and alignment.

CSCL~\cite{yoo2025cscl} introduced curriculum learning with synthetic code-switching data and demonstrated improved bilingual transfer between English and a target language. 
Similarly, SynCS~\cite{wang2025syncs} investigated synthetic code-switching during multilingual pretraining and showed improved cross-lingual alignment across several languages. 
However, SynCS constructs code-switching data only from bilingual language pairs between English and each target language, rather than jointly mixing three or more languages within the same example.

Earlier work such as CoSDA-ML~\cite{qin2020cosdaml} explored multilingual code-switching data augmentation for fine-tuning mBERT on zero-shot cross-lingual NLP tasks.
Our work focuses on the empirical evaluation of code-switching involving three or more languages within individual examples during LLM instruction tuning, a setting that remains underexplored.
While prior studies include token-, phrase-, and sentence-level code-switching strategies, we focus on simple sentence-level mixing that is easy to construct yet remains effective for multilingual instruction tuning.

\section{Multilingual Code-Switching}
We first generate English instruction data, translate the generated instruction examples into multiple languages, and then construct multilingual sentence-level CSD for instruction tuning.

\subsection{Multilingual Instruction Data Construction}
We use Dolma v1.6~\cite{soldaini2024dolma}, a large-scale open pretraining corpus, as the source of English text data for instruction generation, following the setup of CSCL~\cite{yoo2025cscl}.
We sample 50K English texts from Dolma while excluding the code-related \textit{The Stack} subset.

Using \texttt{gpt-4o-mini}~\cite{openai2024gpt4omini}, chosen for its favorable balance between cost and generation quality, we generate multiple-choice instruction data from sampled English texts.
For each instruction example, we generate a multiple-choice question-answering pair consisting of a context snippet selected from the text, a question, four answer options, and the correct answer.
We generate 3--5 instruction examples for each text.
The prompt used for instruction data generation is provided in Table~\ref{sup-tab:instruction_prompt}.

Through this process, we construct approximately 200K English instruction examples comprising roughly 25M tokens.
The resulting instruction examples contain 127.6 tokens on average, with lengths ranging from 57 to 586 tokens.
Figure~\ref{fig:csd_example}(a) illustrates an example of the generated English instruction data.

We then translate the generated instruction examples into Japanese (JA), Korean (KO), and Chinese (ZH) using \texttt{gpt-4o-mini}, resulting in parallel multilingual instruction data across four languages, including English (EN).
The prompt used for translation is provided in Table~\ref{sup-tab:translation_prompt}.

\subsection{Sentence-Level Code-Switching Data}
Following prior work on synthetic code-switching data (CSD)~\cite{yoo2025cscl}, we construct sentence-level multilingual CSD for instruction tuning. 
Unlike prior bilingual settings that mainly focus on transfer between English and a target language, our setting jointly mixes three or more languages within the same instruction example.

We generate sentence-level multilingual CSD by selecting and combining sentence units from parallel instruction examples in different languages. 
We construct each context by mixing sentences from different language versions of the same example. 
The question and answer options are also selected from different language versions of the same example.
To maintain answer consistency and readability, all answer options within a single example are written in the same language.

To determine the language order, we cycle through all permutations of the selected languages across examples: $3! = 6$ permutations in the trilingual setting and $4! = 24$ permutations in the quadrilingual setting.
Within each example, we repeatedly follow the assigned language order for the context sentences, the question, and the answer options, treating the entire set of answer options as a single unit and preserving the original sentence order.

Figures~\ref{fig:csd_example}(b) and (c) show examples of bilingual and multilingual sentence-level CSD, respectively.
Sentence-level CSD is simpler to construct than more fine-grained token-level code-switching while remaining effective for multilingual instruction tuning.

\begin{table*}[t]
    \centering
        \begin{adjustbox}{width=0.9\linewidth}
        \begin{tabular}{llccccc}
            \toprule
            Setting & Method & EN & JA & KO & ZH & Avg. \\
            \midrule
            Baseline & Gemma-2-2B
            & $34.4$ & $27.4$
            & $29.7$ & $27.8$
            & $29.8$ \\
            \midrule
            \multirow{3}{*}{\textbf{E+J}}
            & Concat
            & $82.8 \pm 2.8$ & $69.9 \pm 2.3$
            & $69.4 \pm 1.9$ & $75.5 \pm 3.3$
            & $74.4 \pm 2.6$ \\
            & CSD
            & $83.4 \pm 0.9$ & $72.6 \pm 1.4$
            & $71.5 \pm 0.4$ & $77.4 \pm 1.0$
            & $76.2 \pm 0.9$ \\
            & $\Delta$
            & \hl{$+0.7 \pm 2.1$} & \hl{$+2.7 \pm 1.2$}
            & \hl{$+2.0 \pm 1.6$} & \hl{$+1.9 \pm 2.4$}
            & \hl{$+1.8 \pm 1.7$} \\
            \midrule
            \multirow{3}{*}{\textbf{E+K}}
            & Concat
            & $81.4 \pm 0.8$ & $70.6 \pm 1.2$
            & $70.6 \pm 1.0$ & $76.2 \pm 1.1$
            & $74.7 \pm 1.0$ \\
            & CSD
            & $83.6 \pm 1.1$ & $72.4 \pm 0.9$
            & $72.1 \pm 1.1$ & $77.3 \pm 1.4$
            & $76.4 \pm 0.9$ \\
            & $\Delta$
            & \hl{$+2.3 \pm 1.8$} & \hl{$+1.9 \pm 1.4$}
            & \hl{$+1.5 \pm 1.8$} & \hl{$+1.1 \pm 2.5$}
            & \hl{$+1.7 \pm 1.8$} \\
            \midrule
            \multirow{3}{*}{\textbf{E+Z}}
            & Concat
            & $83.7 \pm 1.5$ & $70.5 \pm 1.2$
            & $71.2 \pm 0.4$ & $77.5 \pm 1.4$
            & $75.7 \pm 1.0$ \\
            & CSD
            & $84.3 \pm 1.5$ & $71.0 \pm 0.9$
            & $71.1 \pm 0.8$ & $78.1 \pm 0.8$
            & $76.2 \pm 0.5$ \\
            & $\Delta$
            & \hl{$+0.5 \pm 0.4$} & \hl{$+0.5 \pm 1.9$}
            & $0.0 \pm 0.9$ & \hl{$+0.7 \pm 1.4$}
            & \hl{$+0.4 \pm 0.9$} \\
            \bottomrule
        \end{tabular}%
        \end{adjustbox}
    \vspace{-2mm}
    \caption{Results of bilingual training settings
    on Belebele for Gemma-2-2B
    (mean $\pm$ standard deviation over three runs).
    Concat denotes multilingual concatenation without code-switching,
    and CSD denotes code-switching data.
    $\Delta$ denotes the performance difference between CSD and Concat
    (CSD $-$ Concat).
    Positive improvements are highlighted in \hl{green}.
    E, J, K, and Z denote English, Japanese, Korean, and Chinese, respectively.}
    \label{tab:gemma2_bi_results}
\end{table*}

\begin{table*}[t]
    \centering
        \begin{adjustbox}{width=0.9\linewidth}
        \begin{tabular}{llccccc}
            \toprule
            Setting & Method & EN & JA & KO & ZH & Avg. \\
            \midrule
            Baseline & Gemma-2-2B
            & $34.4$ & $27.4$
            & $29.7$ & $27.8$
            & $29.8$ \\
            \midrule
            \multirow{3}{*}{\textbf{E+J+K}}
            & Concat
            & $80.6 \pm 3.5$ & $70.3 \pm 2.4$
            & $69.2 \pm 2.2$ & $73.8 \pm 3.3$
            & $73.5 \pm 2.8$ \\
            & CSD
            & $84.0 \pm 1.3$ & $73.3 \pm 1.4$
            & $72.5 \pm 0.8$ & $78.0 \pm 1.2$
            & $77.0 \pm 1.1$ \\
            & $\Delta$
            & \hl{$+3.4 \pm 2.2$} & \hl{$+3.0 \pm 1.4$}
            & \hl{$+3.2 \pm 1.4$} & \hl{$+4.2 \pm 2.2$}
            & \hl{$+3.5 \pm 1.8$} \\
            \midrule
            \multirow{3}{*}{\textbf{E+J+Z}}
            & Concat
            & $81.9 \pm 2.0$ & $70.3 \pm 2.7$
            & $69.8 \pm 1.9$ & $77.1 \pm 1.4$
            & $74.8 \pm 1.9$ \\
            & CSD
            & $83.4 \pm 1.1$ & $72.7 \pm 0.6$
            & $72.9 \pm 0.5$ & $78.5 \pm 1.2$
            & $76.9 \pm 0.4$ \\
            & $\Delta$
            & \hl{$+1.5 \pm 3.0$} & \hl{$+2.4 \pm 2.5$}
            & \hl{$+3.2 \pm 2.2$} & \hl{$+1.4 \pm 2.4$}
            & \hl{$+2.1 \pm 2.3$} \\
            \midrule
            \multirow{3}{*}{\textbf{E+K+Z}}
            & Concat
            & $79.0 \pm 1.8$ & $67.6 \pm 1.6$
            & $67.1 \pm 1.2$ & $73.4 \pm 0.8$
            & $71.8 \pm 1.3$ \\
            & CSD
            & $83.9 \pm 0.2$ & $72.9 \pm 0.3$
            & $72.4 \pm 0.9$ & $78.5 \pm 0.3$
            & $76.9 \pm 0.2$ \\
            & $\Delta$
            & \hl{$+4.9 \pm 1.7$} & \hl{$+5.3 \pm 1.7$}
            & \hl{$+5.3 \pm 2.1$} & \hl{$+5.1 \pm 0.7$}
            & \hl{$+5.2 \pm 1.5$} \\
            \midrule
            \multirow{3}{*}{\textbf{E+J+K+Z}}
            & Concat
            & $80.1 \pm 1.0$ & $70.1 \pm 1.4$
            & $68.7 \pm 0.7$ & $75.3 \pm 1.1$
            & $73.5 \pm 0.4$ \\
            & CSD
            & $85.0 \pm 1.0$ & $74.7 \pm 0.2$
            & $73.0 \pm 0.4$ & $79.7 \pm 1.2$
            & $78.1 \pm 0.2$ \\
            & $\Delta$
            & \hl{$+4.9 \pm 1.4$} & \hl{$+4.7 \pm 1.5$}
            & \hl{$+4.3 \pm 0.4$} & \hl{$+4.4 \pm 1.3$}
            & \hl{$+4.6 \pm 0.5$} \\
            \bottomrule
        \end{tabular}%
        \end{adjustbox}
    \vspace{-2mm}
    \caption{Results of multilingual training settings on Belebele.
    E, J, K, and Z denote English, Japanese, Korean, and Chinese, respectively.}
    \label{tab:multilingual_results}
\end{table*}

\begin{table*}[t]
    \centering
    \begin{adjustbox}{width=0.9\linewidth}
    \begin{tabular}{llccccc}
        \toprule
        Setting & Method & EN & JA & KO & ZH & Avg. \\
        \midrule
        Baseline & Gemma-2-2B
        & $29.2$ & $24.3$ & $24.4$ & $24.0$ & $25.5$ \\
        \midrule
        \multirow{3}{*}{\textbf{E+J+K}}
        
        & Concat
        & $50.8 \pm 2.0$ & $39.5 \pm 1.3$
        & $37.8 \pm 0.7$ & $41.0 \pm 1.8$
        & $42.3 \pm 1.5$ \\
        & CSD
        & $52.2 \pm 0.4$ & $40.5 \pm 0.2$
        & $38.9 \pm 0.2$ & $42.3 \pm 0.3$
        & $43.5 \pm 0.3$ \\
        & $\Delta$
        & \hl{$+1.4 \pm 1.5$} & \hl{$+1.0 \pm 1.2$}
        & \hl{$+1.1 \pm 0.6$} & \hl{$+1.2 \pm 1.5$}
        & \hl{$+1.2 \pm 1.2$} \\
        \midrule
        \multirow{3}{*}{\textbf{E+J+Z}}
        & Concat
        & $52.0 \pm 0.6$ & $40.3 \pm 0.4$
        & $38.5 \pm 0.3$ & $42.1 \pm 0.3$
        & $43.2 \pm 0.2$ \\
        & CSD
        & $52.8 \pm 0.5$ & $40.8 \pm 0.5$
        & $39.6 \pm 0.2$ & $43.0 \pm 0.3$
        & $44.1 \pm 0.2$ \\
        & $\Delta$
        & \hl{$+0.8 \pm 0.3$} & \hl{$+0.4 \pm 0.7$}
        & \hl{$+1.1 \pm 0.2$} & \hl{$+0.9 \pm 0.1$}
        & \hl{$+0.8 \pm 0.2$} \\
        \midrule
        \multirow{3}{*}{\textbf{E+K+Z}}
        & Concat
        & $49.7 \pm 0.6$ & $38.3 \pm 1.0$
        & $37.3 \pm 0.9$ & $40.3 \pm 0.9$
        & $41.4 \pm 0.9$ \\
        & CSD
        & $52.2 \pm 0.4$ & $40.4 \pm 0.1$
        & $39.3 \pm 0.2$ & $42.4 \pm 0.4$
        & $43.6 \pm 0.2$ \\
        & $\Delta$
        & \hl{$+2.5 \pm 0.4$} & \hl{$+2.1 \pm 0.9$}
        & \hl{$+1.9 \pm 1.0$} & \hl{$+2.1 \pm 0.5$}
        & \hl{$+2.2 \pm 0.7$} \\
        \midrule
        \multirow{3}{*}{\textbf{E+J+K+Z}}
        & Concat
        & $50.5 \pm 0.8$ & $39.6 \pm 0.9$
        & $37.8 \pm 0.9$ & $41.2 \pm 1.1$
        & $42.3 \pm 0.9$ \\
        & CSD
        & $52.7 \pm 0.2$ & $40.9 \pm 0.4$
        & $39.4 \pm 0.3$ & $42.8 \pm 0.3$
        & $44.0 \pm 0.2$ \\
        & $\Delta$
        & \hl{$+2.2 \pm 0.7$} & \hl{$+1.3 \pm 0.7$}
        & \hl{$+1.7 \pm 0.7$} & \hl{$+1.7 \pm 0.9$}
        & \hl{$+1.7 \pm 0.7$} \\
        \bottomrule
    \end{tabular}
    \end{adjustbox}
    \vspace{-2mm}
    \caption{Results of multilingual training settings on MMLU.}
    \label{tab:gemma2_mmlu_results}
\end{table*}

\section{Experiments and Analysis}
\subsection{Implementation Details}\label{subsec:imple}
\PAR{Model Setup}
We use the pretrained Gemma-2-2B~\cite{team2024gemma} as our main model.
We additionally evaluate the pretrained Qwen2-1.5B~\cite{yang2024qwen2technicalreport} in trilingual and quadrilingual settings on Belebele, with results reported in Table~\ref{tab:qwen2_multilingual_results} in the appendix.
Our training setup is based on the supervised fine-tuning (SFT) implementation in Hugging Face's TRL~\cite{vonwerra2020trl} library.
We train for one epoch with a learning rate of $5\times10^{-5}$, batch size 32, and a cosine learning rate scheduler. 
We employ parameter-efficient fine-tuning using LoRA~\cite{hu2022lora} with rank 16.

\PAR{Computational Resources and Training Time}\\
\noindent All experiments are conducted using four NVIDIA A100 GPUs with a batch size of 32. 
Each model is trained for one epoch, which requires approximately one hour.

\PAR{Comparison Methods}
We compare multilingual CSD with concatenation baselines without code-switching to analyze the effectiveness of multilingual code-switching during instruction tuning.
We construct bilingual, trilingual, and quadrilingual versions of CSD that always include English: (EN-JA, EN-KO, EN-ZH), (EN-JA-KO, EN-JA-ZH, EN-KO-ZH), and (EN-JA-KO-ZH). 
For comparison, we also construct concatenation baselines without code-switching.
For controlled comparison, all settings use the same number of training examples (200K); only the way in which languages are combined differs.
For example, in the EN-JA-KO-ZH concatenation baseline, the first 50K samples are in English, followed by 50K samples each in Japanese, Korean, and Chinese.

\PAR{Evaluation Data}
For evaluation, we use Belebele~\cite{bandarkar2024belebele}, a multilingual multiple-choice reading comprehension benchmark in which the same questions are translated into multiple languages by experts fluent in English and the target language.
We use 900 parallel test samples across EN, JA, KO, and ZH, enabling direct multilingual comparison across the four languages. 
An example of the Belebele test data is shown in Figure~\ref{sup-fig:belebele_example}.
We additionally evaluate our method on MMLU~\cite{hendrycks2021measuring} across EN, JA, KO, and ZH.
We use \texttt{cais/mmlu}\footnote{\url{https://huggingface.co/datasets/cais/mmlu}} for English and the professionally translated versions from \texttt{openai/MMMLU}\footnote{\url{https://huggingface.co/datasets/openai/MMMLU}} for Japanese, Korean, and Chinese.
We use the full test set of 14,042 questions per language.

\PAR{Evaluation Metric}
For each fine-tuned setting, we perform three runs with different random seeds. We compute accuracy (\%) on the full evaluation set for each run and report the mean and standard deviation across the three runs.
Avg. denotes the arithmetic mean of accuracy across the four evaluation languages.
Following the EleutherAI LM Evaluation Harness~\cite{biderman2024lessons,eval-harness}, we adopt likelihood-based multiple-choice evaluation instead of free-form generation for more stable multilingual evaluation.
We compute the conditional likelihood of each answer choice and select the choice with the highest likelihood as the prediction.

\subsection{Results on Bilingual Settings}
Table~\ref{tab:gemma2_bi_results} presents the results of bilingual training settings on Belebele. 
Overall, CSD consistently improves average performance compared to concatenation baselines across all bilingual settings.

English, Japanese, and Chinese improve across all bilingual settings.
Korean improves in the E+J and E+K settings and remains nearly unchanged in the E+Z setting.
These results suggest that code-switching is more effective than simple concatenation for multilingual transfer.

These trends are consistent with prior studies on bilingual CSD. 
In particular, prior work such as CSCL and SynCS also reported improved multilingual transfer in EN+KO and EN+ZH settings, respectively.
However, unlike prior work that mainly investigated bilingual CSD during large-scale pretraining settings (e.g., 1B tokens in CSCL and a 60B-English/600M-Chinese-token pretraining corpus with up to 2B additional synthetic code-switching tokens), our results show that bilingual CSD is also effective during instruction tuning with substantially smaller training data (approximately 25M tokens in the English source data).

\subsection{Results on Multilingual Settings}
Tables~\ref{tab:multilingual_results} and~\ref{tab:gemma2_mmlu_results} present the results of trilingual and quadrilingual training settings on Belebele and MMLU, respectively.
Compared to bilingual settings, multilingual CSD remains effective even when three or more languages are mixed within the same instruction example. 

On Belebele, the quadrilingual setting (E+J+K+Z) achieves the highest average accuracy of 78.1\%, improving over its concatenation baseline by 4.6 percentage points.
Japanese improves by 4.7 percentage points in this setting.
The largest average improvement occurs in the trilingual E+K+Z setting (+5.2 percentage points).
These results suggest that the benefits of code-switching can extend beyond bilingual settings, even when three or more languages are jointly mixed within the same training example.

On MMLU, CSD improves average accuracy across all trilingual and quadrilingual settings, with gains ranging from 0.8 to 2.2 percentage points.
The E+K+Z setting shows the largest average improvement, while the quadrilingual setting gains 1.7 percentage points.
These results indicate that increasing the number of training languages does not necessarily yield larger improvements.

\section{Conclusion}
In this work, we investigated multilingual code-switching data (CSD) for multilingual instruction tuning across English, Japanese, Korean, and Chinese. 
Our experiments with Gemma-2-2B showed that multilingual sentence-level CSD consistently improves average performance across the four evaluation languages in bilingual, trilingual, and quadrilingual settings on Belebele, and in trilingual and quadrilingual settings on MMLU.
The results further suggest that multilingual code-switching remains effective beyond bilingual transfer settings.

Additional experiments with Qwen2-1.5B (Appendix~\ref{app:qwen_results}) show smaller and less consistent gains, suggesting that the effectiveness of multilingual CSD depends on the underlying model and language combination.
In addition, our findings show that simple sentence-level code-switching can provide consistent multilingual improvements during instruction tuning. 
We hope these findings will inform future work on multilingual instruction tuning and multilingual code-switching research.

\section*{Limitations}
Our experiments focus on sentence-level code-switching data (CSD) across English and three East Asian languages (Chinese, Japanese, and Korean). 
Although we evaluate on both Belebele and MMLU, our evaluation is limited to multiple-choice benchmarks.
Whether the observed improvements extend to open-ended multilingual generation remains an open question.
Future work should investigate more fine-grained token- or phrase-level CSD, larger-scale training datasets, and more diverse multilingual settings covering broader language families and lower-resource languages.

\section*{Acknowledgments}
This work used the UTokyo Azure computing service~\cite{MakotoNakamura20250030}.

\bibliography{custom}

\newcommand\beginsupplement{%
        \setcounter{table}{0}
        \renewcommand{\thetable}{\Alph{table}}%
        \setcounter{figure}{0}
        \renewcommand{\thefigure}{\Alph{figure}}%
     }
\beginsupplement
\appendix
\section{Additional Results on Qwen2-1.5B}
\label{app:qwen_results}

Table~\ref{tab:qwen2_multilingual_results} reports the results of trilingual and quadrilingual instruction tuning on Belebele for Qwen2-1.5B.
For Qwen2-1.5B, CSD improves average accuracy on Belebele in three of the four multilingual settings, with the largest gain of 0.8 percentage points in the quadrilingual setting.
The E+J+Z setting shows a slight decrease of 0.1 percentage points.
These results suggest that the benefits extend beyond Gemma-2-2B, although their magnitude and consistency depend on the model and language combination.

\begin{table*}[t]
    \centering
    \begin{adjustbox}{width=0.9\linewidth}
    \begin{tabular}{llccccc}
        \toprule
        Setting & Method & EN & JA & KO & ZH & Avg. \\
        \midrule
        Baseline & Qwen2-1.5B & 29.1 & 31.0 & 28.6 & 30.7 & 29.9 \\
        \midrule
        \multirow{3}{*}{\textbf{E+J+K}}
        & Concat
        & $81.5 \pm 1.1$ & $67.1 \pm 0.9$
        & $68.6 \pm 0.4$ & $78.1 \pm 0.8$
        & $73.8 \pm 0.6$ \\
        & CSD
        & $81.7 \pm 0.7$ & $68.6 \pm 0.6$
        & $68.6 \pm 0.6$ & $78.5 \pm 0.8$
        & $74.3 \pm 0.7$ \\
        & $\Delta$
        & \hl{$+0.2 \pm 0.5$} & \hl{$+1.5 \pm 0.3$}
        & $+0.0 \pm 0.5$ & \hl{$+0.4 \pm 0.2$}
        & \hl{$+0.5 \pm 0.1$} \\
        \midrule
        \multirow{3}{*}{\textbf{E+J+Z}}
        & Concat
        & $81.8 \pm 0.6$ & $68.2 \pm 0.6$
        & $68.2 \pm 0.2$ & $79.5 \pm 0.3$
        & $74.4 \pm 0.3$ \\
        & CSD
        & $81.9 \pm 0.9$ & $68.4 \pm 0.8$
        & $68.1 \pm 1.0$ & $79.0 \pm 0.6$
        & $74.4 \pm 0.7$ \\
        & $\Delta$
        & \hl{$+0.1 \pm 0.4$} & \hl{$+0.2 \pm 1.1$}
        & $-0.1 \pm 0.9$ & $-0.6 \pm 0.6$
        & $-0.1 \pm 0.6$ \\
        \midrule
        \multirow{3}{*}{\textbf{E+K+Z}}
        & Concat
        & $82.0 \pm 1.0$ & $67.3 \pm 0.8$
        & $67.4 \pm 0.8$ & $78.9 \pm 0.8$
        & $73.9 \pm 0.7$ \\
        & CSD
        & $82.0 \pm 0.2$ & $68.1 \pm 0.8$
        & $68.4 \pm 0.8$ & $79.2 \pm 0.2$
        & $74.4 \pm 0.5$ \\
        & $\Delta$
        & $+0.0 \pm 0.8$ & \hl{$+0.7 \pm 0.5$}
        & \hl{$+1.0 \pm 1.4$} & \hl{$+0.3 \pm 1.0$}
        & \hl{$+0.5 \pm 0.8$} \\
        \midrule
        \multirow{3}{*}{\textbf{E+J+K+Z}}
        & Concat
        & $81.6 \pm 0.2$ & $68.0 \pm 0.3$
        & $67.8 \pm 0.5$ & $78.6 \pm 0.5$
        & $74.0 \pm 0.1$ \\
        & CSD
        & $81.8 \pm 0.7$ & $69.3 \pm 1.7$
        & $68.7 \pm 1.0$ & $79.4 \pm 0.4$
        & $74.8 \pm 0.8$ \\
        & $\Delta$
        & \hl{$+0.2 \pm 0.6$} & \hl{$+1.3 \pm 1.9$}
        & \hl{$+0.9 \pm 0.6$} & \hl{$+0.8 \pm 0.3$}
        & \hl{$+0.8 \pm 0.7$} \\
        \bottomrule
    \end{tabular}
    \end{adjustbox}
    \vspace{-2mm}
    \caption{Results of multilingual training settings on Belebele
    for Qwen2-1.5B (mean $\pm$ standard deviation over three runs).
    }
    \label{tab:qwen2_multilingual_results}
\end{table*}

\begin{figure*}[t]
    \centering
    \includegraphics[width=0.95\linewidth]{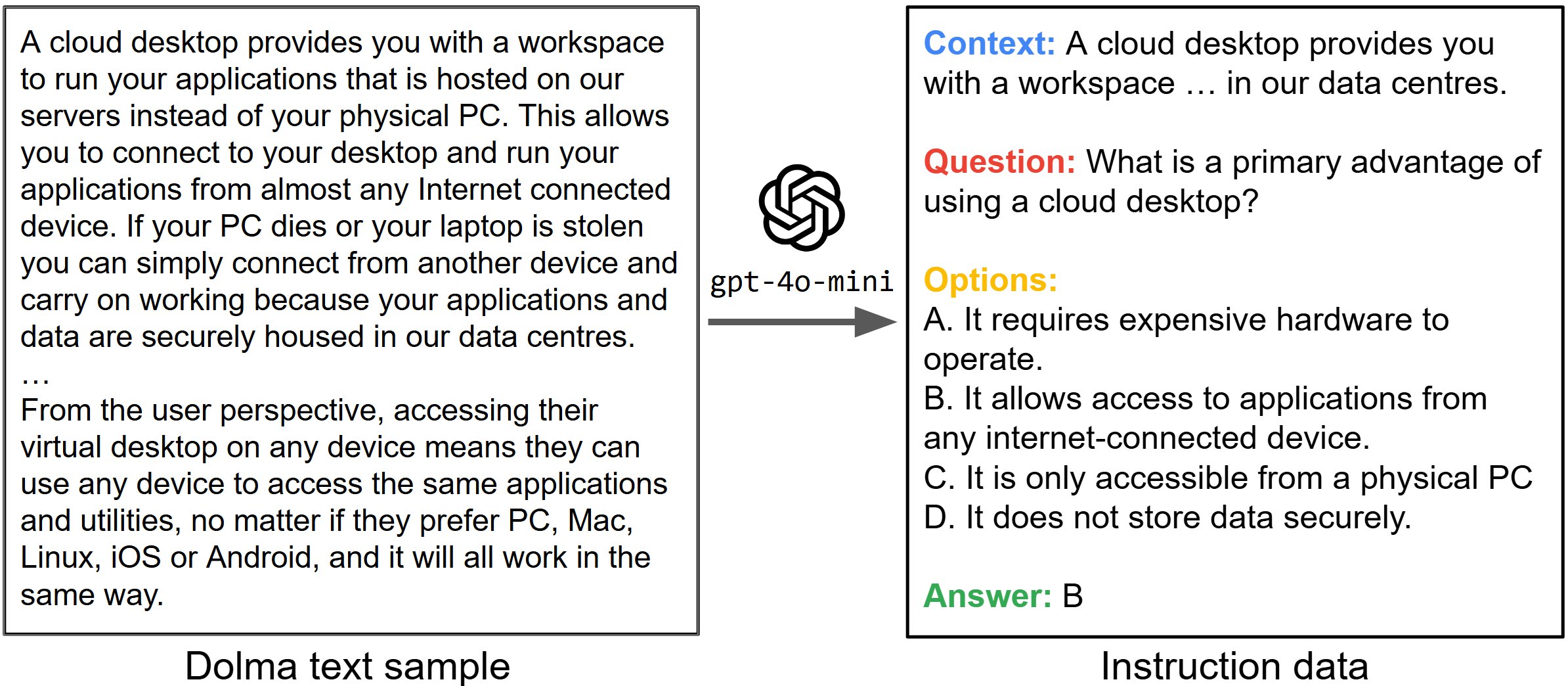}
    \vspace{-2mm}
    \caption{Example of instruction data generation from a Dolma text sample using GPT-4o-mini.}
    \label{sup-fig:dolma_instruction}
\end{figure*}

\section{Data Examples}
\subsection{Instruction Data Generation}
Figure~\ref{sup-fig:dolma_instruction} shows an example of instruction data generation from a Dolma~\cite{soldaini2024dolma} text sample. 
Given a raw text passage, \texttt{gpt-4o-mini}~\cite{openai2024gpt4omini} generates a context snippet selected from the text, a question, candidate options, and the corresponding correct answer in a multiple-choice question-answering format. 
This process enables the automatic construction of instruction-style data from unlabeled text resources.

\subsection{Manual Translation Quality Check}
Before creating code-switching and concatenation datasets, we manually inspected 20 translations by comparing them with their English sources.
We checked whether each translation preserves the source meaning and whether the correct answer remains valid given the translated context, question, and answer options.
This process serves as a qualitative sanity check rather than a quantitative assessment of translation quality.

\subsection{Multilingual Code-Switching Data}
\begin{figure*}[t]
    \centering
    \includegraphics[width=0.95\linewidth]{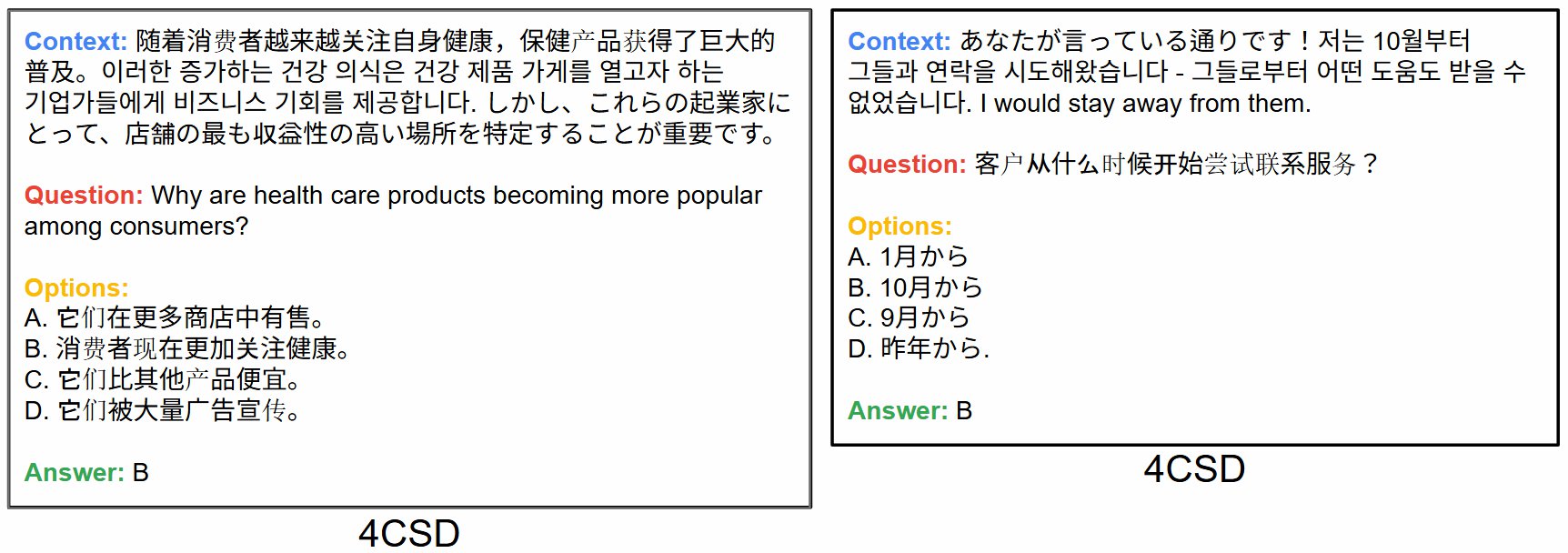}
    \vspace{-2mm}
    \caption{Examples of multilingual sentence-level code-switching data (CSD) used in our work. 
    The examples jointly mix English, Japanese, Korean, and Chinese within the same instruction example while preserving the original question-answer structure.}
    \label{sup-fig:multilingual_csd_example}
\end{figure*}

Figure~\ref{sup-fig:multilingual_csd_example} illustrates examples of multilingual sentence-level CSD used in our work. 
Each example jointly mixes English, Japanese, Korean, and Chinese within the same instruction example while preserving the original question-answer structure. 
These examples demonstrate how multilingual instruction examples are constructed from parallel multilingual data.

\subsection{Belebele Test Data}
\begin{figure*}[t]
    \centering
    \includegraphics[width=0.95\linewidth]{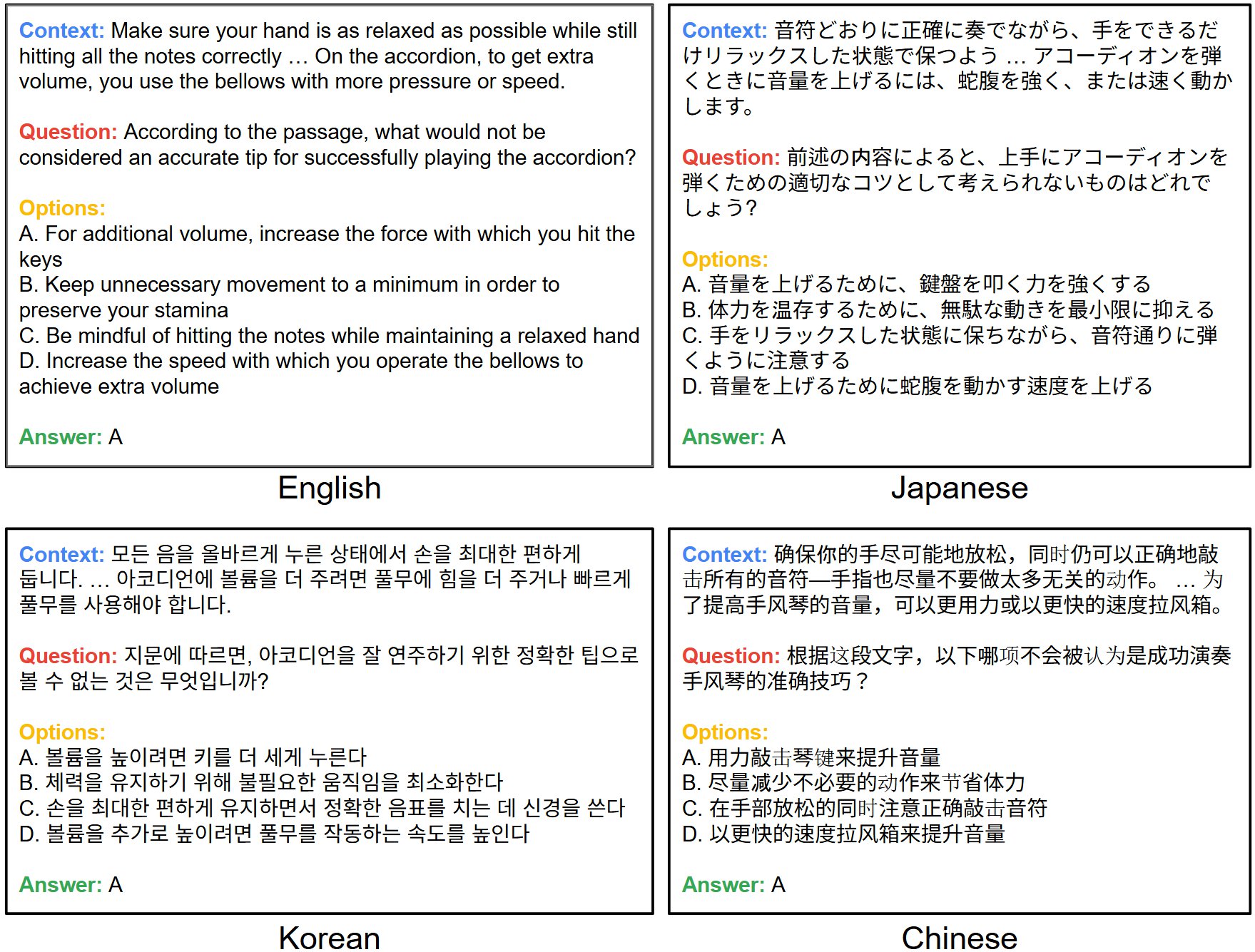}
    \vspace{-2mm}
    \caption{Example of a parallel test sample from the Belebele benchmark. 
    The same reading comprehension example is provided in English, Japanese, Korean, and Chinese.}
    \label{sup-fig:belebele_example}
\end{figure*}

Figure~\ref{sup-fig:belebele_example} presents an example from the Belebele benchmark~\cite{bandarkar2024belebele} used as test data in our experiments. 
Belebele is a multilingual reading comprehension benchmark consisting of parallel question-answer pairs across multiple languages. 
The figure shows one example of the same question presented in the four languages, demonstrating the parallel structure of the benchmark and enabling direct multilingual comparison.

\begin{figure*}[t]
    \centering
    \includegraphics[width=\textwidth]{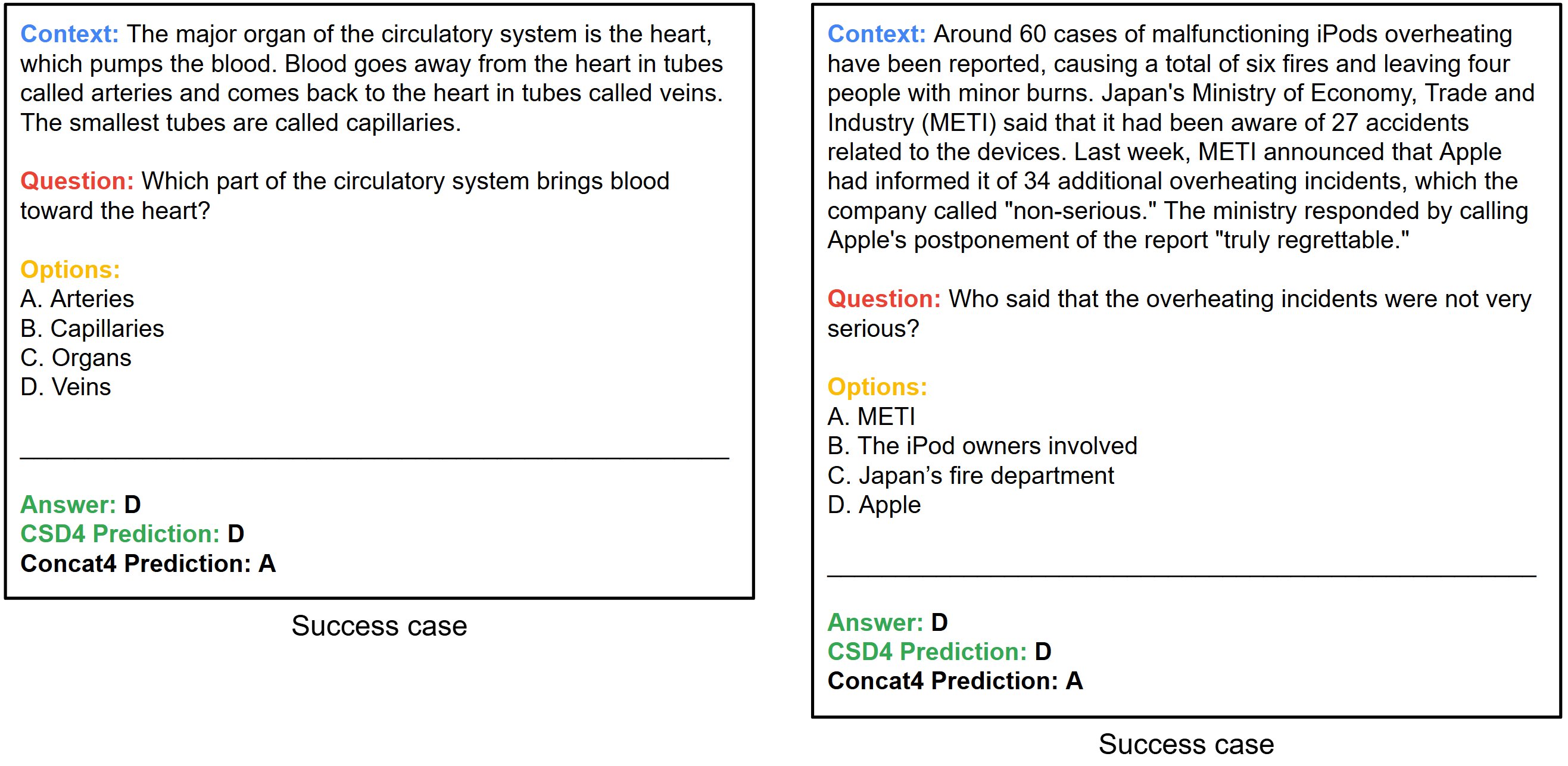}
    \caption{
    Examples where multilingual CSD correctly predicts the answer while the concatenation baseline fails. 
    CSD4 and Concat4 denote the quadrilingual multilingual CSD and concatenation settings, respectively.
    }
    \label{sup-fig:csd_success_case}
\end{figure*}

\begin{figure*}[t]
    \centering
    \includegraphics[width=\textwidth]{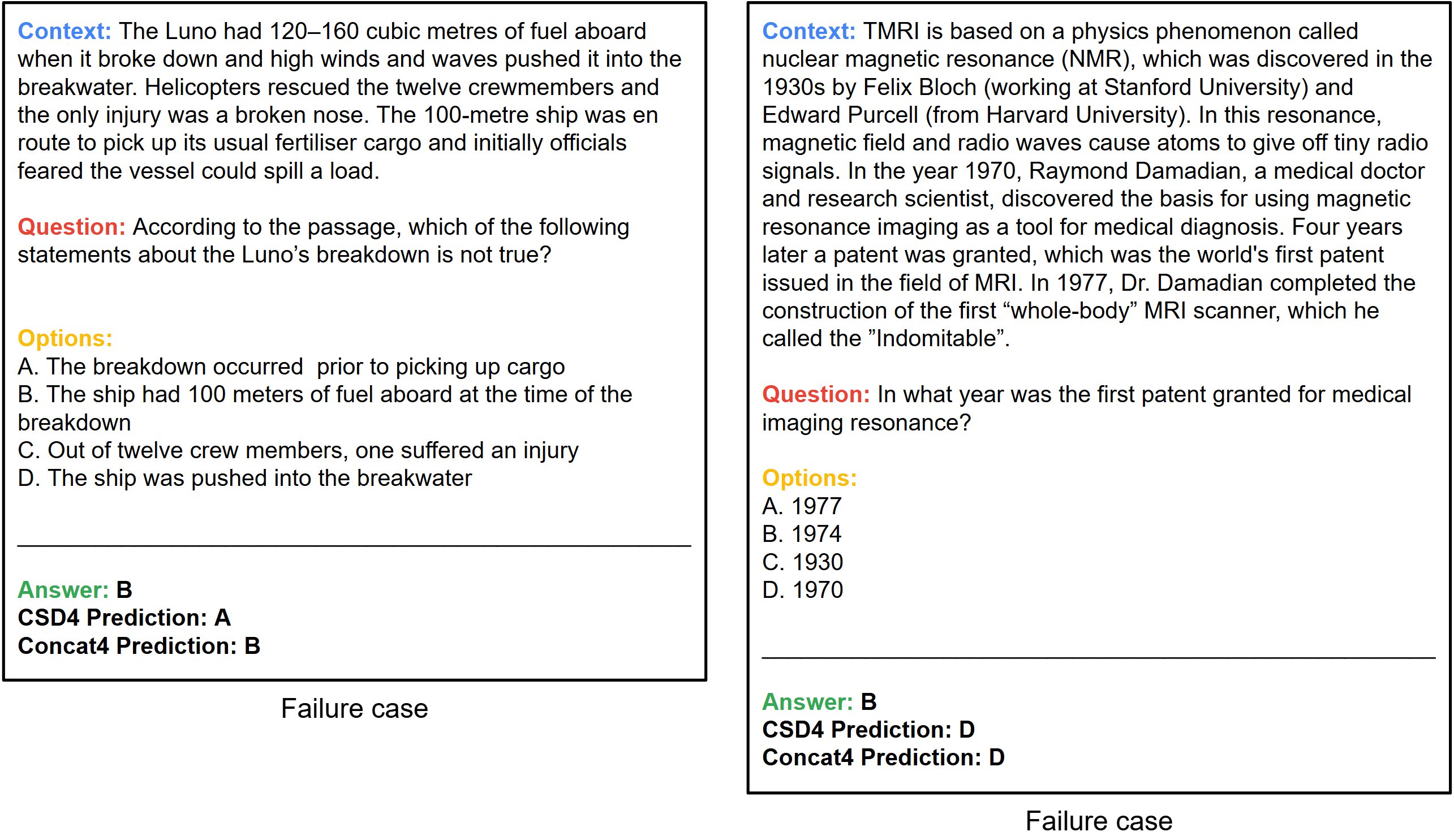}
    \caption{
    Failure cases where multilingual CSD still produces an incorrect prediction. 
    Despite the improvements brought by multilingual CSD, complex multilingual reasoning and contextual understanding remain challenging.
    }
    \label{sup-fig:csd_failure_case}
\end{figure*}

\section{Qualitative Analysis}
Figure~\ref{sup-fig:csd_success_case} shows examples where multilingual CSD correctly predicts the answer while the concatenation baseline fails. 
These examples suggest that multilingual code-switching can improve multilingual understanding by exposing the model to more diverse multilingual contexts during instruction tuning.

Figure~\ref{sup-fig:csd_failure_case} presents failure cases where multilingual CSD still produces an incorrect prediction. 
Despite the improvements brought by multilingual CSD, such cases indicate that multilingual instruction tuning remains challenging, particularly for complex multilingual reasoning scenarios.

\section{Prompts}
We use prompts for two stages: (1) instruction data generation from English Dolma data and (2) translation of the generated instruction data into Japanese, Korean, and Chinese.

First, instruction data is generated from English Dolma data using the prompt shown in Table~\ref{sup-tab:instruction_prompt}. 
The generated English instruction data is then translated into Japanese, Korean, and Chinese using the translation prompt shown in Table~\ref{sup-tab:translation_prompt}.

\begin{table*}[t]
\begin{tcolorbox}[colback=gray!5, colframe=black, 
colbacktitle=gray!30, coltitle=black, title=System message, fonttitle=\bfseries, rounded corners]
You are creating high-quality multiple-choice reading comprehension QA data for instruction tuning of large language models.

Requirements: \\
- Generate 3 to 5 diverse QA pairs from the provided raw text. \\
- For each QA pair, select a relevant context snippet from the raw text. \\
- Each QA pair should be independently understandable using its own context. \\
- Different QA pairs should cover different parts or ideas from the raw text when possible. \\
- Context snippets should typically be around 3 sentences long. \\
- Include enough information to answer the question naturally. \\

Question requirements: \\
- Questions must require reading and understanding the context. \\
- Do NOT use external knowledge. \\
- Avoid trivial or duplicate questions. \\
- Avoid questions answerable from common world knowledge alone. \\
- Prefer natural reading comprehension questions. \\
- Avoid copying sentences directly from the context. \\

Multiple-choice requirements: \\
- Each question must have exactly 4 options: A, B, C, D. \\
- Only one option should be correct. \\
- Incorrect options should be plausible and context-related. \\
- Avoid obviously incorrect or joke options. \\ 
- Keep option styles and lengths reasonably similar. \\

Return ONLY valid JSON. \\ 

Format:
[
  \{
    "context": "...",
    "question": "...",
    "options": \{
      "A": "...",
      "B": "...",
      "C": "...",
      "D": "..."
    \},
    "answer": "<CORRECT\_OPTION>"
  \}
]
\end{tcolorbox}
\begin{tcolorbox}[colback=gray!5, colframe=black, 
colbacktitle=gray!30, coltitle=black, title=User prompt, fonttitle=\bfseries, rounded corners]
Generate 3 to 5 diverse multiple-choice reading comprehension QA pairs from the following raw text. \\
Raw text: \\
\{\{TEXT\}\}
\end{tcolorbox}
\vspace{-1em}
\caption{Prompt used for instruction data generation.}
\label{sup-tab:instruction_prompt}
\end{table*}

\begin{table*}[t]
\begin{tcolorbox}[colback=gray!5, colframe=black, 
colbacktitle=gray!30, coltitle=black, title=System message, fonttitle=\bfseries, rounded corners]
You are a professional translator.\\

Task: \\
Translate English multiple-choice reading comprehension QA data into natural and fluent \{language\}. \\
  
Requirements:  \\
- Translate the values of "context", "question", and all option texts.  \\
- Do NOT translate JSON keys.  \\
- Do NOT translate the answer option letter.  \\
- Keep the JSON structure exactly unchanged.  \\
- Do NOT remove, add, or alter newline characters ("\textbackslash n"). \\
- Preserve all option keys exactly as A, B, C, D.  \\
- Do NOT reorder, remove, merge, or split options.  \\
- Do NOT reorder, remove, merge, or split QA pairs.  \\
- The output array length MUST exactly match the input array length.  \\
- Ensure that the translated options remain consistent with the correct answer letter.  \\
- Preserve factual meaning, numbers, dates, URLs, names, and entities.  \\
- Do NOT use external knowledge.  \\
- Return ONLY valid JSON.  \\
  
Output format:
[
  \{
    "context": "...",
    "question": "...",
    "options": \{
      "A": "...",
      "B": "...",
      "C": "...",
      "D": "..."
    \},
    "answer": "<CORRECT\_OPTION>"
  \}
]
\end{tcolorbox}
\begin{tcolorbox}[colback=gray!5, colframe=black, 
colbacktitle=gray!30, coltitle=black, title=User prompt, fonttitle=\bfseries, rounded corners]
Translate the following English multiple-choice QA data into \{language\}.  \\
Input JSON:  \\
\{\{QA\_JSON\}\}
\end{tcolorbox}
\vspace{-1em}
\caption{Prompt used for translating instruction data into Japanese, Korean, and Chinese.}
\label{sup-tab:translation_prompt}
\end{table*}

\end{document}